# Automated Dynamic Bayesian Networks for Predicting Acute Kidney Injury Before Onset

**David Gordon, MS**[1,2,*], **Panayiotis Petousis, PhD**[3], **Anders O. Garlid, PhD**[2], **Keith Norris, MD, PhD**[4], **Katherine Tuttle, MD**[5], **Susanne B. Nicholas, MD, MPH, PhD**[4,+], and **Alex A.T. Bui, PhD**[1,2,3,+] **on behalf of CURE-CKD**

[1]Department of Bioengineering, University of California, Los Angeles, CA, USA;
[2]Medical & Imaging Informatics (MII) Group, Department of Radiological Sciences, University of California, Los Angeles, CA, USA; [3]UCLA Clinical and Translational Science Institute, Los Angeles, CA, USA; [4]Department of Medicine, David Geffen School of Medicine at UCLA, Los Angeles, CA, USA; [5]Providence Health Care, Spokane, WA, USA

[*]*Correspondence to:* UCLA Medical Imaging Informatics, 924 Westwood Blvd., Suite 420, Los Angeles, CA 90024.
 Email-address: d.gordon@ucla.edu (D. Gordon)
[+]Senior authors had equal contributions

**Abstract**

*Several algorithms for learning the structure of dynamic Bayesian networks (DBNs) require an* a priori *ordering of variables, which influences the determined graph topology. However, it is often unclear how to determine this order if feature importance is unknown, especially as an exhaustive search is usually impractical. In this paper, we introduce Ranking Approaches for Unknown Structures (RAUS), an automated framework to systematically inform variable ordering and learn networks end-to-end. RAUS leverages existing statistical methods (Cramer's V (CV), chi-squared test (Chi²), and information gain (IG)) to compare variable ordering, resultant generated network topologies, and DBN performance. RAUS enables end-users with limited DBN expertise to implement models via command line interface. We evaluate RAUS on the task of predicting impending acute kidney injury (AKI) from inpatient clinical laboratory data. Longitudinal observations from 67,460 patients were collected from our electronic health record (EHR) and Kidney Disease Improving Global Outcomes (KDIGO) criteria were then applied to define AKI events. RAUS learns multiple DBNs simultaneously to predict a future AKI event at different time points (i.e., 24-, 48-, 72-hours in advance of AKI). We also compared the results of the learned AKI prediction models and variable orderings to baseline techniques (logistic regression, random forests, and extreme gradient boosting). The DBNs generated by RAUS achieved 73-83% area under the receiver operating characteristic curve (AUCROC) within 24-hours before AKI; and 71-79% AUCROC within 48-hours before AKI of any stage in a 7-day observation window. CV-DBN and IG-DBN ranked serum albumin and estimated glomerular filtration rate (eGFR) as the top two features within 48-hours before AKI, resulting in higher overall performance. Insights from this automated framework can help efficiently implement and interpret DBNs for clinical decision support.*

## 1 Introduction

Acute kidney injury (AKI) involves a rapid decline in renal function over the span of hours or a few days, resulting in significant sequalae with noted increase in morbidity and mortality.[1] AKI is estimated to occur in 10-15% of all hospitalizations, 22-57% of patients in intensive care units (ICUs), and 12-22% of postoperative patients.[2-5] The estimated inpatient costs related to AKI in the US ranges from $5.4-24.0 billion per year.[6-8] In many cases, AKI may be preventable if suspected sufficiently early.[9] Unfortunately, AKI often goes undetected before it is too late for preventive measures to be initiated and/or effective.[10] In this paper, we explore whether AKI can be identified in a timely manner in inpatient settings, focusing on patients with chronic kidney disease (CKD), a group considered at high-risk for AKI events. Although AKI can occur for many reasons, it is well-established that individuals with CKD have higher odds of an AKI event, and the likelihood of kidney failure increases up to 40 times with underlying CKD.[11] As such, clinical decision support tools detecting the potential onset of AKI in CKD patients are especially critical, helping to preserve remaining kidney function and limit disease progression as well as acute complications from kidney failure.

Several studies have suggested the potential of artificial intelligence and machine learning (ML) to detect AKI before onset (BOS).[10, 12-15] However, these models are largely "static" in estimating the risk of an individual over time, failing to take into consideration evolving information. Here, we assess automated dynamic Bayesian networks (DBNs) for early detection of AKI in CKD patients, combining a trajectory of past and current clinical observations to determine how quickly we can predict impending AKI over a 7-day time horizon in an at-risk CKD population.[16] Notably, prior analysis reveals higher performance of models when predicting AKI on a patient subpopulation with reduced kidney function, suggesting this subgroup is a good fit for practical implementation.[10, 17] We focus on the problem of efficiently identifying viable, interpretation-ready network topologies for the DBN given unknown ranked feature importance. In the absence of *a priori* (or expert) knowledge about the relationships between variables in a domain, the structure of a DBN must be sought using an exhaustive search of potential graph structures, a Monte Carlo Markov Chain (MCMC) simulation, or other method tailored to the target application domain.[18-20] But if a preliminary ordering can be ascertained, greedy search methods, like the K2 algorithm,[21] can establish a directed acyclic graph (DAG) for the DBN intra-structure. To this end, we introduce *Ranking Approaches for Unknown Structures* (RAUS),[22] a general automated framework that uses multiple filter ranking approaches (e.g., Cramer's V, chi-squared, and information gain) for DBN structure learning algorithms, thus guiding variable ordering when it is unknown (which we informally refer to as using unknown structure learning experts). This process facilitates exploration of competing approaches, providing interpretation-ready output for review including automatically generated network graphs that illustrate feature importance for better model explainability.[23]

Using a dataset of >67,000 hospitalized patients with CKD extracted from our institution's electronic health record (EHR), we used RAUS to generate and evaluate DBNs for predicting AKI events over three different time windows (24-, 48-, and 72-hours) based on the Kidney Disease Improving Global Outcomes (KDIGO) criteria, thus requiring the implementation of multiple DBNs simultaneously. We compare RAUS' results to embedded feature ranking methods derived from other data-driven models (logistic regression, random forests, and extreme gradient boosting), describe the resultant DBN performance, and discuss insights gained from applying this automated framework to a real-world dataset.

## 2 Background and Related Work

### 2.1 Predicting Acute Kidney Injury Events

Several efforts have looked to identify predictors of AKI events. In one inpatient study that reported an area under the receiver operating characteristic curve (AUCROC) of 0.74 using a logistic regression, the reported top ten features predictive of AKI of any stage within 24-hours were: serum creatinine (SCr), blood urea nitrogen (BUN), heart rate, anion gap, BUN/SCr ratio, respiratory rate, glucose, white blood cell count (WBC), potassium, and $O_2$ saturation.[14] In a subsequent study at the same site, the reported top features predictive of a gradient boosted machine model for predicting at least Stage 2 AKI within 48-hours included change in SCr, length of stay, saturation $PaO_2/FiO_2$ ratio, current SCr, change in BUN, current total 12 hour urine output, current serum calcium, current serum phosphate, lowest systolic pressure (24-hours), and highest heart rate within a 24-hour period. This later model achieved comparable performance with an AUCROC of 0.73.[13] Still other predictive models that have examined AKIs report the use of heart rate, respiratory rate, temperature, Glasgow coma scale, features around glomerular filtration rate (GFR) (at admission, mean during admission, changes in GFR, preadmission mean outpatient GFR, etc.), body mass index, alkaline phosphatase, glucose, features around SCr (yearly baseline, 48-hour mean), arterial blood gas pH, and comorbidities. No significant improvements were seen, with AUCROC performance ranging from 0.71-0.76,[12, 24] and PR-AUC ranging from 0.26-0.32.[10] Across these studies, different ML approaches (random forests, gradient boosted forests, and recurrent neural networks) and regression methods (logistic regression) were used. These AKI studies demonstrated variation in feature rankings that can occur for several reasons, such as model selection, data availability, study population, time window, and AKI severity. As such, there remains no clear consensus on the importance of different features involved in AKI prediction, which makes this a challenging clinical decision-making problem. Notably, across these models, varying definitions of AKI were used and none of these described models have been adopted for real-world usage.

### 2.2 Feature Ranking Techniques

When an order is assumed or known for feature importance, heuristic search methods can be used to learn graphical models, including dynamic Bayesian networks, even if the true ordering is unknown.[20] Many ML algorithms and statistical approaches involve determining the relative importance of different features to discriminate between classes or estimate values. Broadly, such feature ranking methods can be categorized as using a filter, wrapper, or embedded algorithms.[25] Filter methods rank features based on the relationship between the predictor(s) and outcome and are computationally inexpensive as they do not require training a model. Examples of filter methods include chi-squared, Cramer's V, Pearson's correlation, analysis of variance (ANOVA), information gain, and Markov blankets, to name a few.[25,26] Wrapper methods examine model performance using subsets of features and iteratively add/remove features to achieve the best performance. As such, wrapper methods are more computationally expensive.[25] Well-known examples of wrapper methods include forward selection and backward elimination. Finally, embedded methods are a hybrid of filter and wrapper methods that rank features as part of the modeling process, and include L1 regularization (LASSO regression), L2 regularization (ridge regression), and decision trees. Many of the AKI predictive models cited above use embedded feature ranking methods to determine a valid ranking for feature selection.[10,12,13,14,24] Although filter methods including chi-squared and information gain have been previously explored with belief networks for efficient feature ranking, they and other methods (e.g., Cramer's V) have not been explored in the context of automated DBNs, which we explore here as part of RAUS.[20]

## 3 Methods

### 3.1 Dataset and Predictive Task Formulation

The data used in this study is a subset of the Providence Health System (PHS) and the University of California Los Angeles Health (UCLA) Center for Kidney Disease Research, Education, and Hope (CURE-CKD) Repository,[27, 28] drawing on those patients seen at UCLA. CURE-CKD captures observational and outcomes data from the EHR on individuals deemed at-risk for CKD (e.g., history of hypertension, diabetes, or prediabetes) or diagnosed with CKD. The dataset represents a total of >2.6 million adults seen over a decade (1/2006-12/2017) without a history of kidney failure treated by dialysis or transplant. For purposes of this study, we searched the repository to identify patients with a diagnosis of CKD and then subsequent inpatient admission at UCLA. This query resulted in 67,460 individuals that were used in model development and testing. Table 1 provides overall statistics about this population, as well as the EHR data elements extracted from CURE-CKD.

**Identifying AKIs: Gold standard definition.** As AKI is not always identified within the medical record (i.e., AKI events can go undocumented and thus not present in terms of diagnostic codes, per ICD-9/10), we used the Kidney Disease Improving Global Outcomes clinical guideline definition[16] to establish a gold standard (Table 2). Patients are defined as experiencing AKI of any stage if they: 1) have an estimated glomerular filtration rate (eGFR) <60 mL/min/1.73m$^2$ and concomitant increase in SCr >1.5 times from admission's baseline (within 7 days); or 2) have a SCr increase of ≥0.3 mg/dL within a 48-hour period. Table 1 also summarizes the results of applying these definitions to our cohort. Temporal lab features considered as part of our model are also listed, with the mean eGFR for these individuals falling within CKD Stage 3b (i.e., moderate CKD, eGFR 45-30 mL/min/1.73m$^2$). Figure 1 illustrates the AKI states changing over a 7-day period ($t_0$-$t_6$). Notably, the diagram shows that a patient may have unresolved AKI and may remain with low eGFR over time.

**Predictive task.** We posed three prediction windows for detecting AKI events in terms of 24-hour periods: 1) within 72-hours before onset; 2) within 48-hours before onset; and 3) within 24-hours before onset. These timepoints were selected to assess how early an AKI event could be identified in advance, providing a way to set timepoints that would also be clinically meaningful within an inpatient setting.

**Data preprocessing.** We represented each CKD patient's data as a sequence of events grouped into up to seven 24-

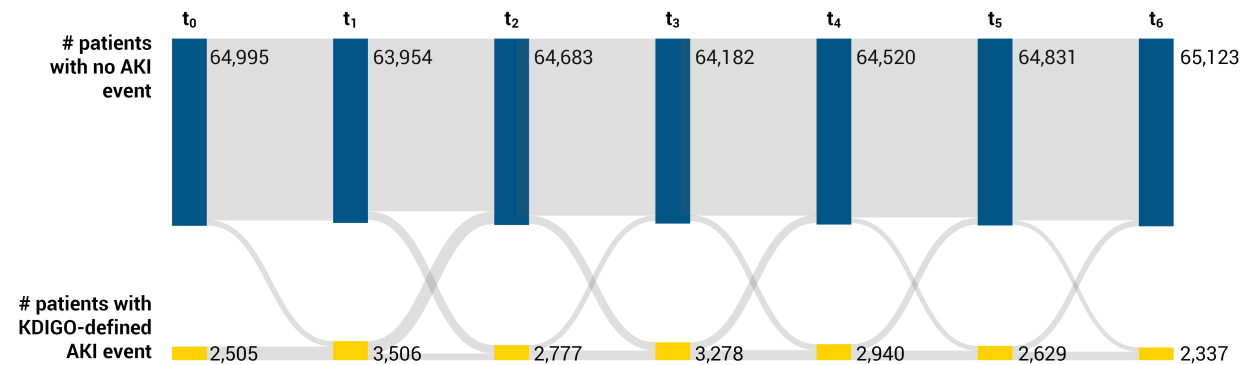

**Figure 1.** Sankey diagram depicting the number of AKI and non-AKI events at each timestep. An individual may enter and leave the AKI state more than once during the 7-day period, resulting in a steady state total number of individuals across the two categories.

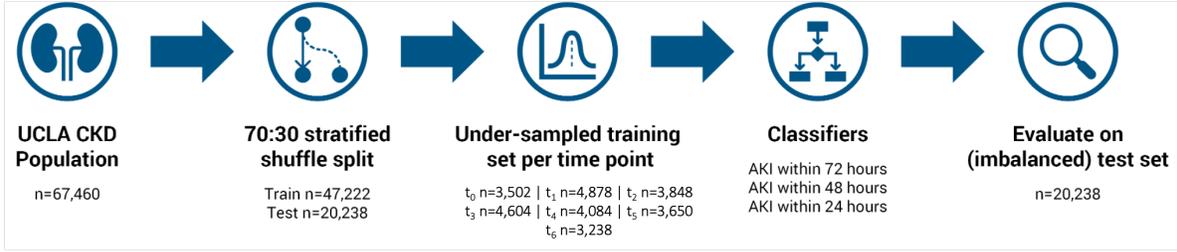

**Figure 2.** Train/test split, showing the performed internal validation as well as the balancing strategy. The majority class (non-AKI) was under-sampled at each timestep, resulting in a training set of n=27,804. A 5-fold cross validation was then conducted.

hour periods during hospitalization. Multiple lab observations of the same type (e.g., several eGFR measures) within a 24-hour period were typically available, so the observation closest to the end of the 24-hour period was selected. We also discretized observations within the population. For eGFR, we discretized values based on established levels for CKD staging, starting at Stage 3a,[29, 30] embedding the severity of the individual's underlying condition into the model. For all other continuous laboratory values, we established bins based on the interquartile ranges (IQRs) (i.e., minimum, 25%, 50%, 75%, maximum). P-values were calculated to understand the individual significance of each variable, and dependent on the specified cut-off, they were removed from consideration. For example, given the relatively large dataset of >67,000 instances, we selected only variables at a more stringent alpha level with p-value <0.01 for input to RAUS.

**Dataset splits.** To train and test the DBNs, we used a holdout procedure through a 70:30 stratified shuffled split for both case (AKI event) and control (non-AKI event) data (Figure 2). We note here that the ratio of patients with any AKI to non-AKI in the dataset is 8,023:59,437, or about 13 individuals with an AKI event for every 100 without an AKI event. To handle this imbalance at each timestep we used under-sampling, randomly selecting the same number of controls as cases to create seven balanced training subsets while still preserving the temporal information for AKI and non-AKI events in subsequent timesteps.

### 3.2 Learning the DBN: RAUS Automated Framework

Figure 3 illustrates RAUS, an automated pipeline for learning and evaluating DBNs using multiple rank learning approaches. Consolidating multiple well-established statistical methods and (D)BN algorithms, five modules are integrated in RAUS, which we briefly describe here:

1. <u>Rank learning</u>. To start, given a set of variables and corresponding data, RAUS generates a list of variables ranked by different metrics. Three measures are then calculated per variable: chi-square ($Chi^2$), a basic measure of sta-tistical significance; Cramer's V (CV),[23] a measurement used to assess the strength of relationship between categorical variables; and information gain (IG),[31] an entropic measure based on observations and targeted (outcome) variables. The intent is to construct, in descending order for each method, a sequence of variables from more to less informative. RAUS can then be configured to select all ranked features, best-k (i.e., top) ranked features, or a top percentile of the ranked features for consideration, which is innovative for DBN implementations. We believe these rank learning approaches are reasonable to initialize variable ordering as a DBN assumes stationarity (i.e., the topological order is assumed to be the same at subsequent timesteps). However, additional ranking methods can be readily added to RAUS.

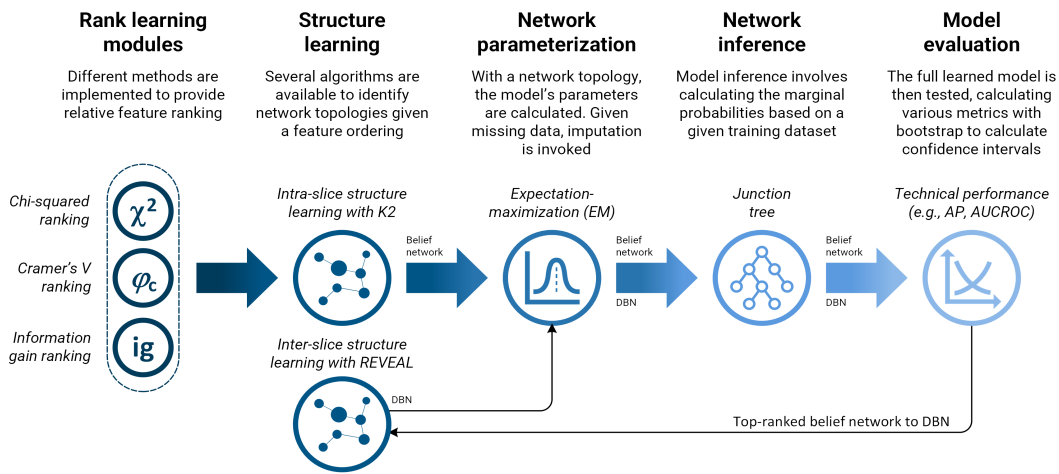

**Figure 3**. RAUS automated framework. Different ranking modules can be configured and compared, resulting first in the initial construction of belief networks that are evaluated. The top model is then selected for usage as part of a learned dynamic Bayesian network.

2. Structure learning. Results from the rank learning module are passed on to structure learning algorithms that then search for DAGs consistent with the given variable ordering. Previous research on belief networks and DBNs have resulted in several methods for inferring intra-slice (capturing variable dependencies within a timestep) and inter-slice (capturing node dependencies between timesteps) relationships.[18, 32] RAUS uses K2 for intra-slice structure learning and REVEAL[33] for inter-slice structure learning. Again, additional structure learning algorithms can be easily added to RAUS, facilitating comparisons.

3. Network parameterization. RAUS then calculates the associated conditional probability tables (CPTs). To handle missing data, a common occurrence with EHR-based datasets, expectation-maximization (EM) is employed. EM is an iterative procedure that uses the current estimate of parameters to find the expectation of complete data and utilize it to find a maximum likelihood estimate of the parameter.[34] Other network parameterization methods can also be added to RAUS for specific purposes.

4. Network inference. The junction tree (JT) algorithm is at the heart of most exact inference techniques, and can be constructed from the DAGs.[35] Given the DAG structures, JT computes marginals on graphs by creating a tree of graphs and carrying out message passing on the tree.[18] Constructing the JT includes moralizing (ensuring a node and its parents are in the same clique) and triangulating the graph (ensuring all loops containing four or more nodes contain a chord to produce elimination orderings). For DBNs, a smoother was used with JT, implementing forward/backward operators.[18] The result of this module are the marginal probabilities. Different inference algorithms can be used in RAUS.

5. Evaluation and model selection. Lastly, given the instantiated set of (D)BNs, we compute performance metrics, including average precision-recall score (AP) and AUCROC. To assess variability, we bootstrap computations based on a user-specified parameter with replacement (e.g., 1,000 re-randomization runs) to compute 95% confidence intervals (CIs). Based on selected technical metrics, RAUS then outputs a ranked list of DBNs.

At each step, RAUS provides output for review, including automatic generation of different (interpretation-ready) visualizations (e.g., DBN and BN network topologies, graphs comparing variable ranking across methods). Note that in the autogenerated network visualizations, feature importance is shown counterclockwise (novel for DBN implementations) starting at the most important feature (i.e., yellow node) with the outcome variable immediately after the most important feature. Source code scripts are provided to facilitate the use of RAUS across different configurations (e.g., running only DBN or BN generation based on all rank learning methods; learning and selecting the top DBN, etc.). RAUS is implemented using Python v3.7.3 as well as python packages Pandas[36], NumPy[37], Scikit-learn[38], Multiprocessing[39], Oct2py[40], Matplotlib[41], SciPy[42], and Networkx.[43] We use GNU Octave v4.2.2 and Bayes Net Toolbox to implement the BNs and DBNs.[44] We also use the R package Amelia II for baseline models that require complete datasets as input.[45] Given the potentially large number of variables and permutations for a single run, parallelization across modules is employed. RAUS can be implemented via command line interface.

The source code for RAUS is available in GitHub: https://github.com/dgrdn08/RAUS.

### 3.3 AKI Model Comparison Experiments

To compare automated DBNs with conventional embedded feature ranking methods, we created three baselines: a logistic regression (LR) model, which uses maximum likelihood estimation (MLE) as its objective function, liblinear solver[46] for its optimization, and L2 regularization; random forest (RF), where the number of estimators/trees in the forest is set to 100 and entropy is used for information gain to measure the quality of a split; and extreme gradient boosting (XGB), where the booster/learner is a tree function, the maximum depth is set to 6, and the importance type is gain.[47] The Cramer's V DBN (CV-DBN), chi-squared DBN (Chi$^2$-DBN), information gain DBN (IG-DBN), LR model, RF model, and XGB model were each evaluated on the holdout imbalanced test set (n=20,238), with model parameters tuned to prefer AP given the highly imbalanced nature of the dataset. However, LR, RF, and XGB require complete datasets as input; therefore, LR, RF, and XGB use an EM imputed dataset,[45] whereas CV-DBN, Chi$^2$-DBN, and IG-DBN take as input an incomplete dataset and then use EM with its learned network topology to infer results.

We also compared the results of the different DBNs' predictive performance with respect to the LR, RF, and XGB models and assessed the DBNs at varying thresholds for precision/recall tolerance. Specifically, we evaluated the DBNs, LR, RF, and XGB models at three precision levels: 40% (i.e., 2 true positives for every 3 false positives) for within 24-hours BOS; 33% (i.e., 1 true positive for every 2 false positives) for within 48-hours BOS; and 20% (i.e., 1 true positive for every 4 false positives) for within 72-hours BOS. The basic idea being that the end-user can personalize the model implementation based on their desired precision level. Further, we examined the case agreement (true positives, false negatives, and false positives) between the three prediction windows (24-, 48-, and 72- hours BOS) for the DBN.

## 4 Results

RAUS was run on the CURE-CKD dataset using an AWS EC2 ubuntu instance with 18 physical cores and 36 virtual CPUs.[48] A total of 12 DBN and 9 BN models were automatically generated and evaluated using this framework. Figure 4 presents a subset of the generated DBN topologies (CV-DBN within 48 BOS, Chi$^2$-DBN within 48 BOS, and IG-DBN within 48 BOS). A comparison of results across methods is provided below.

### 4.1 Variable Ranking Comparison

A total of eight (8) variables were statistically significant based on a cutoff p-value statistic of <0.01. From these variables, Table 3 presents the learned rankings across all six models (CV-DBN, Chi$^2$-DBN, IG-DBN, LR, RF, and XGB) for each of the three prediction windows, comparing the filter- and embedded-based feature importance. Interestingly, there was no consensus across the methods, and there were differences across the time windows. For instance, for within 24-hours BOS, IG-DBN ranked serum albumin as its top feature, whereas CV-DBN and Chi$^2$-DBN ranked eGFR as their top feature. Further, for within 48 and within 72-hours BOS, feature ranking between CV-DBN and IG-DBN were the same for the top three features (one of them being eGFR) whereas Chi$^2$-DBN did not include eGFR in the top three features. Moreover, for RF, potassium and sodium were not ranked in the bottom three features for any prediction window. Notably, the differences in ranking across the 24-, 48-, and 72-hour periods suggests that the relative importance of these variables changes over time, suggesting the need for a dynamic set of models that is used accordingly (vs. a singular model for all time points). This finding also suggests the variation seen in past literature is dependent both on the prediction window and underlying severity of the AKI event.

### 4.2 AKI DBN Comparisons

**Model structures.** Unsurprisingly, the use of different algorithms for both ranking and structure learning result in variations in DBN structures. Figure 4 shows differences, for example, between CV-DBN, Chi$^2$-DBN, and IG-DBN for the same 48-hour BOS prediction window. Notably, CV-DBN and IG-DBN rank eGFR as the second most important feature, with a higher number of outdegrees (6, one edge to each descendent/child) in comparison to Chi$^2$-DBN, which ranks calcium as the second most important feature, resulting in a lower number of outdegrees (4 descendants).

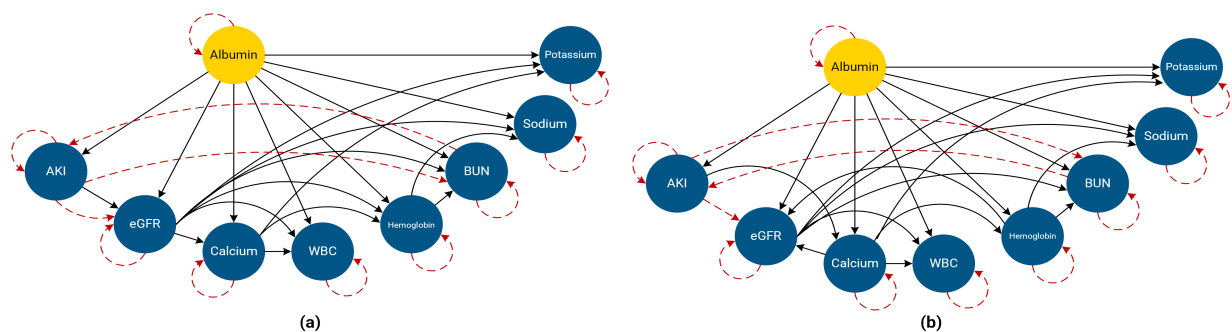

(a)   (b)

**Figure 4.** Selected compact representations of DBNs learned for detecting the onset of an AKI event, 48-hours before onset. Solid arrows indicate intra-time slice dependency in the DBN between variables; red dashed arrows specify inter-slice relationships. **(a)** CV-DBN and IG-DBN resulted in the same network structures (though the Hemoglobin node was more important than the WBC node for IG-DBN); **(b)** Chi$^2$-DBN (note that we modified the variable order in the visualization to match (a) for easier comparison (the order of feature importance was Albumin, Calcium, Hemoglobin, eGFR, WBC, BUN, Sodium, and Potassium). While several expected relationships are repeated across these learned structures – notably, inter-slice relationships are identical – differences exist in the intra-slice structures. Abbreviations: (BUN) blood urea nitrogen; (WBC) white blood cell count.

**Model performance.** Table 4 demonstrates the performance of using K2 to establish intra-slice topologies and REVEAL to establish inter-slice topologies for all rank-learning approaches. For the within 24-hour prediction window, CV-DBN, $Chi^2$-DBN, and IG-DBN perform similarly and achieve between 0.16-0.39, 0.15-0.38, and 0.15-0.40 AP, respectively. For within 48-hours, IG-DBN and CV-DBN outperform $Chi^2$-DBN and achieve between 0.14-0.25 and 0.13-0.25 AP, respectively. For within 72-hours, IG-DBN and CV-DBN again outperform $Chi^2$-DBN and achieve between 0.10-0.13 AP and 0.10-0.12 AP, respectively.

In comparison, Supplementary Table **S1** describes the performance when using only the K2 algorithm, assuming a more "static" perspective. IG-BN achieves the top AP for the within 24-hours BOS (0.204) and within 48-hours (0.129) prediction windows. But CV-BN instead achieves the top AP for within 72-hours (0.098).

Combining approaches, Supplementary Table **S2** shows the results of using the best K2 models from Supplementary Table **S1** to establish inter-slice topologies using REVEAL. Performance remains on par with Table 4: for within 24-hours BOS, IG-DBN achieves between 0.15-0.40 AP; for within 48-hours, IG-DBN achieves between 0.13-0.25 AP; and for within 72-hours, CV-DBN has 0.10-0.12 AP. In summary, Table 4 shows that using IG or CV variable rankings perform better than $Chi^2$ variable rankings for within 48-hours BOS and within 72-hours BOS prediction windows, while Supplementary Table **S1** shows that IG variable rankings perform the best for most of the prediction windows. As such, we further explored the IG-DBNs in a more in-depth error analysis.

Table 5 summarizes the model comparison test results. IG-DBN outperforms the baseline models in terms of AP score in the within 24-hour prediction window, whereas CV-DBN outperforms the baseline models in terms of AP score in the within 48-hour prediction window. LR outperforms the baseline models in terms of AP score in the within 72-hours BOS prediction window. The variation between models suggests and the underlying (non)linearity further emphasizes the evolutionary observations and processes underlying AKIs. Overall, we found that DBNs using efficient filter-based variable rankings outperformed LR, RF, and XGB models using embedded-based variable rankings for the within 24-hours BOS and within 48-hours BOS prediction windows.

Table 6 presents model performance at a 40% precision level for within 24-hours BOS at $t_5$. IG-DBN improves recall by 48.1%, 42.7%, 32.2%, 11.2%, and 10.8% compared to LR, XGB, RF, CV-DBN, and $Chi^2$-DBN, respectively. Note that for within 24-hours BOS, the closest RF precision level gets to 40% is 21.3%. Table 7 presents model performance at a 33% precision level (i.e., one true positive for every two false positives) for within 48-hours BOS at $t_4$. IG-DBN improves recall by 10.3%, 17.1%, and 29.8% compared to XGB, RF, and LR, respectively. Within 48-hours BOS, the closest RF and XGB precision levels get to 33% is 16.4% and 17.4%, respectively. Table 8 shows model performance at a 20% precision level (i.e., 1 true positive for every 4 false positives) for within 72-hours BOS at $t_3$. IG-DBN improves recall by 16.1%, 5.7%, and 3.9% compared to $Chi^2$-DBN, LR, and CV-DBN, respectively. Note that for within 72-hours BOS, the closest RF and XGB precision levels get to 20% is 11.2% and 10.7%, respectively. Lastly, we note that IG-DBN achieved the best recall for within 24-hours BOS and within 72-hours BOS prediction windows at 40% precision level and 20% precision level, respectively. Similarly, we found that IG-DBN and $Chi^2$-DBN achieve the best recall for within 48-hours BOS prediction window at a 33% precision level.

**Error analysis.** To appreciate if there were subsets of individuals with AKI events that were either consistently correctly "detected" or "missed" over our models over time (i.e., were there individuals, irrespective of time window that were easily detected in advance vs. individuals who we could not identify regardless of time window), we compared the different models. Supplementary Fig. **S1** compares results for the prediction windows for IG-DBN at a 40% precision level within 24-hours BOS, 33% precision level within 48-hours BOS, and 20% precision level for within 72-hours BOS; summary comparisons are given in Supplementary Table **S3**. Supplementary Table **S4** shows the temporal feature distributions for the true positive (TP), false negative (FN), and false positive (FP) analyses for within 24 BOS ∩ within 48 BOS ∩ within 72 BOS for IG-DBN. Note, models were only trained on temporal features; therefore, Supplementary Table **S4** may reveal insight into biases the model learns from the data. For TP analyses, eGFR, BUN, and calcium were found to have significantly different distributions for within 24-hours BOS and within 48-hours BOS; while eGFR, BUN, and albumin were found to have significantly different distributions for within 72-hours BOS. For FN analyses, eGFR, BUN, and albumin were found to have significantly different distributions for within 24-hours BOS, within 48-hours BOS, and within 72-hours BOS. Potassium and calcium were found to have significantly different distributions for within 72-hours BOS FN analyses. For FP analyses, eGFR, BUN, calcium, and albumin were found to have significantly different distributions for within 24-hours BOS, within 48-hours BOS, and within 72-hours. Additionally, hemoglobin was found to have significantly different distributions for within 72-hours FP analyses.

Supplementary Table **S5** shows the static feature distributions for the TP, FN, and FP analyses for within 24 BOS ∩

within 48 BOS ∩ within 72 BOS for IG-DBN. For TP analyses, hypertension was found to have significantly different distributions. For FN analyses, age, gender, race, and hypertension were found to have significantly different distributions. For FP analyses, age and hypertension were found to have significantly different distributions. Lastly, Supplementary Table **S6** shows early true positives (ETPs, those that are detected earlier than noted onset) for within 48-hours BOS and within 72-hours BOS for IG-DBN. 35% of the FN at within 48-hours BOS are TP at within 24-hours BOS; 25% of the FN at within 72-hours BOS are TP at within 48-hours BOS; and 44% of the FN at within 72-hours BOS are TP at within 24-hours BOS.

## 5 Discussion

Using RAUS enabled us to automatically learn variable rankings for DBN development by using Cramer's V, chi-squared, and information gain methods. Notably, by having belief networks compete, we were able to systematically select the best efficient variable ranking method for learning a DBN. Further, we were able to easily compare the generated learned structures and review model performance, which otherwise can be a time-consuming process. In this specific study, we demonstrated the use of RAUS for AKI detection; however, RAUS is a generalized framework that can be applied to numerous problems. Existing open-source software for learning DBNs with unknown structures tend to have a steep learning curve and are not automated, which can make it difficult for end-users to readily implement DBNs. Here, we provide RAUS as an open-source software to help reduce the challenges with implementing DBNs via automating the process end-to-end.

**Understanding AKI.** Our experiments show that the developed CV-DBN, Chi$^2$-DBN, and IG-DBN outperforms comparable LR, RF, and XGB models in terms of AP in predicting an AKI event within 24- and 48-hours BOS. Chi$^2$-DBN ranked eGFR outside the top three features in the within 48-hours BOS and within 72-hours BOS prediction windows, which resulted in lower overall performance. Similarly, RF did not rank potassium and sodium in the bottom three features or hemoglobin in the top three features in all prediction windows, which resulted in lower performance. Yet our performance improvement in CV-DBN, Chi$^2$-DBN, and IG-DBN comes with some cost: while the topology provides a rich representation of the dependencies between AKI risk factors within and across timesteps, it is harder to learn when partially observed.

For the within 24-hours BOS prediction window, we observed a drop in AP at $t_1$ in CV-DBN, Chi$^2$-DBN, and IG-DBN. Further analysis shows a steep decline in recurring AKI events (71% at $t_1$ to 44% at $t_2$) and an influx of new AKI events (56%) (Figure 1). As our models use observations from $t_1$ to predict future AKI events in $t_2$ and $t_2$ represented mostly new AKI events, the computed priors may not have been as helpful at this timestep and could have impacted performance at $t_1$. However, recurring AKI events increase at subsequent timesteps for the within 24-hours prediction window. Similarly, for the within 48-hours prediction window, we observed a lower AP at $t_0$ and $t_1$ in CV-DBN, Chi$^2$-DBN, and IG-DBN. As our models use observations from $t_0$ and $t_1$ to predict future AKI events in $t_2$ and $t_3$, and $t_2$ and $t_3$ represented mostly new AKI events from $t_0$ and $t_1$, the computed priors may have affected the performance at $t_0$ and $t_1$, respectively. Therefore, for the within 48-hours prediction window, we suggest using it after $t_1$ to realize the benefit of the computed priors and leverage the longer lookahead for patients with unresolved AKI events.

**Limitations.** While providing insight into the practical implementation of RAUS, this study has several limitations that we briefly discuss here. First, despite having a large cohort of individuals, the dataset is limited to a single institution and an external validation would be useful. Second, given the complexity of the temporal analysis, we limited this study to a singular 7-day hospitalization window anchored at the point of admission, rather than a "moving" 7-day window – which while providing more event observations would also confound our primary goal of assessing the utility of the framework in this domain. Third, given the considerable heterogeneity in defining the baseline SCr, we used the most common time frame, SCr measured at admission; however, an earlier measurement (e.g., within last 3 months) may also be valid.[49] Finally, we focused on a CKD subpopulation as this is a high risk group for AKI; however, exploring a more general population may be useful.

Given our initial experience and promising results using RAUS, we plan to utilize RAUS on other datasets with different clinical outcomes, such as rapid decline in eGFR (e.g., 40% decline in eGFR per year), as well as across multiple sites. Additional future work will also explore relaxing the topological order assumption, mixture of (unknown) structure learning experts, knowledge graph embeddings, as well as novel imputation methods, such as graph representation learning, which can also take advantage of relationships captured in network topologies.[50, 51]

## 6 Conclusion

RAUS enabled us to automatically learn DBNs to explore a challenging clinical problem across multiple prediction windows simultaneously. We implemented RAUS via command line-interface, which conveniently stored the results and figures in folders for easy downstream interpretation. By predicting AKI before onset, RAUS may be used to help advance clinical decision support by alerting physicians to the need for intervention, particularly in high-risk populations. Further, end-users can personalize the model implementations by adjusting for their desired precision level. We introduced the RAUS open-source software, an automated framework that uses Cramer's V, chi-squared, and information gain to determine variable ordering for the DBN structure learning algorithms when the variable order was unknown. In this study, we investigated the impact of variable order and the use of temporal models on the downstream prediction of impending AKI events in a CKD inpatient setting and demonstrated that using temporal models with automated variable ordering, when accompanied with the appropriate imputation method, improved model performance and scalability. Six machine learning algorithms, CV-DBN, Chi$^2$-DBN, IG-DBN, LR, RF, and XGB were developed with three prediction windows using the EHR. We demonstrated that the use of RAUS (CV-DBN, Chi$^2$-DBN, and IG-DBN) automatically learns DBNs and outperformed static models (LR, RF, and XGB). Specifically, CV-DBN, Chi$^2$-DBN, and IG-DBN outperformed LR, RF, and XGB in terms of AP within 24- and 48-hours BOS while maintaining acceptable levels of precision (40% within 24-hours BOS and 33% within 48-hours BOS).

**Data Availability**

The data used in this study is available from the CURE-CKD program on request at https://www.uclahealth.org/programs/cure-ckd/contact-us.

**Additional Information**

The authors declare that there is no conflict of interest related to this paper.

**Author Contributions**

Most intellectual work was done by DG with guidance from AB. DG was responsible for planning and implementing the paper methodologies. DG was responsible for developing the software, running the experiments, and evaluating the models. PP, AG, KN, KT, SN, and AB reviewed drafts of the paper and provided feedback.

**Acknowledgements**

This work was supported in part by the NIH T32 EB016640, NIH R01 MD014712, and NSF NRT-HDR 1829071. The authors would like to thank Davina Zamanzadeh for her contributions to this paper.

**Table 1.** UCLA population characteristics for individuals with CKD and an inpatient admission between 1/2006-12/2017. Demographics and lab features considered as part of this study are shown. Abbreviations: (AKI) acute kidney injury; (CVD) cardiovascular disease; (eGFR) estimated glomerular filtration rate.

| Population characteristics | | Lab features | Mean (SD) |
| --- | --- | --- | --- |
| Observation period | 2006-2017 | eGFR | 38.0 (15.66) |
| Total patients | 67,460 | Serum calcium (mg/dL) | 9.1 (0.84) |
| Age range (mean/SD) | 20-100 (69/16) | Serum albumin (g/dL) | 3.9 (0.70) |
| Female | 50.2% | White blood cell count (x$10^3$/mL) | 9.6 (7.98) |
| White | 71.2% | Blood urea nitrogen (ml/dL) | 31.3 (19.38) |
| Black | 8.5% | Hemoglobin (g/dL) | 11.9 (2.26) |
| % subjects with an AKI | 11.9% | Serum potassium (mmol/L) | 4.4 (0.70) |
| % subjects with CVD | 55.0% | Serum sodium (mmol/L) | 137.8 (5.06) |
| % subjects with diabetes | 27.0% | | |

**Table 2.** Ground truth labels for AKI of any stage among CKD population using KDIGO criteria.

| Baseline eGFR | Change in Serum Creatinine (SCr) | Diagnosis |
| --- | --- | --- |
| <60 mL/min/1.73 m$^2$ | Increase in SCr to >1.5× baseline, which is known or presumed to have occurred within the past 7 days; or increase in SCr by ≥0.3 mg/dL within 48-hours | AKI+CKD |
| <60 mL/min/1.73 m$^2$ | Increase in SCr to <1.5× baseline, which is known or presumed to have occurred within the past 7 days; or increase in SCr by <0.3 mg/dL within 48-hours | CKD |

**Table 3.** Variable rankings for DBNs (filter-based feature ranking) and baseline comparison models (embedded-based feature ranking). Cramer's V, chi-squared, and information gain were used to determine the input variable ordering (most-to-least important) used in the unknown structure learning for DBNs (assumes stationary, i.e., same ranking across timesteps). Logistic regression, random forest, and extreme gradient boosting models are used as static baselines (non-stationary, i.e., ranking is at final timestep). Abbreviations: (BUN) blood urea nitrogen; (WBC) white blood cell count.

|  |  | Filter-based Ranks | | | Embedded-based Ranks | | |
|---|---|---|---|---|---|---|---|
|  | **Variables** | CV Rank | Chi$^2$ Rank | IG Rank | LR Rank | RF Rank | XGB Rank |
| Within 24-hours | eGFR | 1 | 1 | 2 | 1 | 1 | 1 |
| | BUN | 6 | 6 | 6 | 2 | 2 | 2 |
| | Calcium | 3 | 2 | 3 | 8 | 6 | 8 |
| | Albumin | 2 | 3 | 1 | 4 | 8 | 7 |
| | Hemoglobin | 4 | 4 | 5 | 3 | 7 | 3 |
| | WBC | 5 | 5 | 4 | 5 | 5 | 6 |
| | Sodium | 7 | 7 | 7 | 7 | 4 | 4 |
| | Potassium | 8 | 8 | 8 | 6 | 3 | 5 |
| Within 48-hours | eGFR | 2 | 4 | 2 | 1 | 1 | 1 |
| | BUN | 6 | 6 | 6 | 2 | 4 | 2 |
| | Calcium | 3 | 2 | 3 | 6 | 6 | 7 |
| | Albumin | 1 | 1 | 1 | 4 | 8 | 5 |
| | Hemoglobin | 5 | 3 | 4 | 3 | 7 | 3 |
| | WBC | 4 | 5 | 5 | 5 | 3 | 4 |
| | Sodium | 7 | 7 | 7 | 7 | 5 | 6 |
| | Potassium | 8 | 8 | 8 | 8 | 2 | 8 |
| Within 72-hours | eGFR | 2 | 6 | 2 | 1 | 1 | 1 |
| | BUN | 6 | 5 | 5 | 3 | 5 | 2 |
| | Calcium | 3 | 2 | 3 | 5 | 7 | 8 |
| | Albumin | 1 | 1 | 1 | 2 | 8 | 4 |
| | Hemoglobin | 4 | 3 | 4 | 4 | 6 | 3 |
| | WBC | 5 | 4 | 6 | 6 | 3 | 6 |
| | Sodium | 7 | 7 | 7 | 8 | 4 | 7 |
| | Potassium | 8 | 8 | 8 | 7 | 2 | 5 |

**Table 4.** DBN AUCROC with 95% CIs and average precision-recall score.

| | Models | Metrics | $t_0$ | $t_1$ | $t_2$ |
|---|---|---|---|---|---|
| **Within 24-hours** | CV-DBN | AUCROC CI | 0.815 [0.802, 0.828] | 0.736 [0.715, 0.757] | 0.802 [0.784, 0.821] |
| | | AP | 0.203 | 0.155 | 0.329 |
| | $Chi^2$-DBN | AUCROC CI | 0.828 [0.815, 0.840] | 0.733 [0.712, 0.753] | 0.797 [0.779, 0.816] |
| | | AP | 0.218 | 0.147 | 0.325 |
| | IG-DBN | AUCROC CI | 0.821 [0.807, 0.833] | 0.759 [0.741, 0.777] | 0.823 [0.807, 0.839] |
| | | AP | 0.209 | 0.150 | 0.369 |
| **Within 48-hours** | CV-DBN | AUCROC CI | 0.719 [0.701, 0.740] | 0.730 [0.714, 0.747] | 0.785 [0.767, 0.802] |
| | | AP | 0.134 | 0.141 | 0.258 |
| | $Chi^2$-DBN | AUCROC CI | 0.713 [0.695, 0.731] | 0.726 [0.710, 0.742] | 0.780 [0.763, 0.797] |
| | | AP | 0.088 | 0.126 | 0.222 |
| | IG-DBN | AUCROC CI | 0.722 [0.704, 0.742] | 0.737 [0.719, 0.753] | 0.785 [0.767, 0.801] |
| | | AP | 0.137 | 0.142 | 0.253 |
| **Within 72-hours** | CV-DBN | AUCROC CI | 0.699 [0.683, 0.716] | 0.725 [0.708, 0.742] | 0.754 [0.736, 0.771] |
| | | AP | 0.096 | 0.119 | 0.118 |
| | $Chi^2$-DBN | AUCROC CI | 0.708 [0.691, 0.724] | 0.733 [0.716, 0.749] | 0.743 [0.725, 0.760] |
| | | AP | 0.103 | 0.107 | 0.109 |
| | IG-DBN | AUCROC CI | 0.699 [0.682, 0.716] | 0.732 [0.715, 0.749] | 0.758 [0.741, 0.775] |
| | | AP | 0.099 | 0.119 | 0.123 |
| | **Models** | **Metrics** | $t_3$ | $t_4$ | $t_5$ |
| **Within 24-hours** | CV-DBN | AUCROC CI | 0.826 [0.807, 0.845] | 0.821 [0.799, 0.841] | 0.833 [0.811, 0.854] |
| | | AP | 0.387 | 0.349 | 0.338 |
| | $Chi^2$-DBN | AUCROC CI | 0.825 [0.806, 0.843] | 0.823 [0.802, 0.844] | 0.833 [0.812, 0.853] |
| | | AP | 0.384 | 0.350 | 0.342 |
| | IG-DBN | AUCROC CI | 0.830 [0.813, 0.848] | 0.827 [0.808, 0.848] | 0.828 [0.807, 0.847] |
| | | AP | 0.404 | 0.366 | 0.363 |
| **Within 48-hours** | CV-DBN | AUCROC CI | 0.778 [0.758, 0.797] | 0.761 [0.738, 0.783] | x |
| | | AP | 0.245 | 0.229 | x |
| | $Chi^2$-DBN | AUCROC CI | 0.776 [0.756, 0.795] | 0.763 [0.741, 0.784] | x |
| | | AP | 0.227 | 0.215 | x |
| | IG-DBN | AUCROC CI | 0.779 [0.758, 0.799] | 0.765 [0.742, 0.787] | x |
| | | AP | 0.247 | 0.226 | x |
| **Within 72-hours** | CV-DBN | AUCROC CI | 0.769 [0.751, 0.787] | x | x |
| | | AP | 0.124 | x | x |
| | $Chi^2$-DBN | AUCROC CI | 0.748 [0.728, 0.767] | x | x |
| | | AP | 0.103 | x | x |
| | IG-DBN | AUCROC CI | 0.768 [0.751, 0.788] | x | x |
| | | AP | 0.129 | x | x |

**Table 5.** Model comparisons for CV-DBN, Chi$^2$-DBN, IG-DBN, RF, LR, and XGB at final timestep.

|  | Timestep | Metrics | CV-DBN | Chi$^2$-DBN | IG-DBN | RF | LR | XGB |
|---|---|---|---|---|---|---|---|---|
| **Within 24-hours** | $t_5$ | AUCROC | 0.833 | 0.833 | 0.828 | 0.773 | 0.848 | 0.822 |
|  |  | AP | 0.338 | 0.342 | 0.363 | 0.130 | 0.196 | 0.172 |
| **Within 48-hours** | $t_4$ | AUCROC | 0.761 | 0.763 | 0.765 | 0.707 | 0.795 | 0.758 |
|  |  | AP | 0.229 | 0.215 | 0.226 | 0.096 | 0.149 | 0.110 |
| **Within 72-hours** | $t_3$ | AUCROC | 0.769 | 0.748 | 0.768 | 0.664 | 0.778 | 0.734 |
|  |  | AP | 0.124 | 0.103 | 0.129 | 0.075 | 0.135 | 0.083 |

**Table 6.** Adjusting precision/recall performance to compare the models at a 40% precision level (2 true positives for every 3 false positives) for the within 24-hours BOS prediction window at $t_0$. Confusion matrices are depicted (TP, true positive; FP, false positive; TN, true negative; FN, false negative; P, precision; R, recall; TNR, true negative rate; NPV, negative predictive value; FNR, false negative rate). [1]Closest precision level gets to 40%.

| Model | Within 24-Hours BOS AKI at $t_5$ | | |
|---|---|---|---|
| CV-DBN (Threshold 0.458) | 273 (TP) | 445 (FN) | 0.380 (R) |
| | 403 (FP) | 19117 (TN) | 0.979 (TNR) |
| | 0.404 (P) | 0.977 (NPV) | 0.620 (FNR) |
| Chi²-DBN (Threshold 0.450) | 276 (TP) | 442 (FN) | 0.384 (R) |
| | 407 (FP) | 19113 (TN) | 0.979 (TNR) |
| | 0.404 (P) | 0.977 (NPV) | 0.616 (FNR) |
| IG-DBN (Threshold 0.410) | 353 (TP) | 365 (FN) | 0.492 (R) |
| | 519 (FP) | 19001 (TN) | 0.973 (TNR) |
| | 0.405 (P) | 0.981 (NPV) | 0.508 (FNR) |
| RF (Threshold 0.68) | 122 (TP) | 596 (FN) | 0.170 (R) |
| | 451 (FP) | 19069 (TN) | 0.977 (TNR) |
| | 0.213 (P)[1] | 0.970 (NPV) | 0.830 (FNR) |
| LR (Threshold 0.793) | 8 (TP) | 710 (FN) | 0.011 (R) |
| | 14 (FP) | 19506 (TN) | 0.999 (TNR) |
| | 0.364 (P)[1] | 0.965 (NPV) | 0.989 (FNR) |
| XGB (Threshold 0.780) | 47 (TP) | 671 (FN) | 0.065 (R) |
| | 98 (FP) | 19422 (TN) | 0.995 (TNR) |
| | 0.324 (P)[1] | 0.967 (NPV) | 0.935 (FNR) |

**Table 7.** Adjusting precision/recall performance to compare the models at a 33% precision level (1 true positive for every 2 false positives) for the within 48-hours BOS prediction window at $t_0$. Confusion matrices are depicted (TP, true positive; FP, false positive; TN, true negative; FN, false negative; P, precision; R, recall; TNR, true negative rate; NPV, negative predictive value; FNR, false negative rate). [1]Closest precision level gets to 33%.

| Model | Within 48-Hours BOS AKI at $t_4$ | | |
|---|---|---|---|
| CV-DBN (Threshold 0.394) | 210 (TP) | 508 (FN) | 0.292 (R) |
| | 417 (FP) | 19103 (TN) | 0.979 (TNR) |
| | 0.335 (P) | 0.974 (NPV) | 0.708 (FNR) |
| Chi²-DBN (Threshold 0.397) | 220 (TP) | 498 (FN) | 0.306 (R) |
| | 441 (FP) | 19079 (TN) | 0.977 (TNR) |
| | 0.333 (P) | 0.975 (NPV) | 0.694 (FNR) |
| IG-DBN (Threshold 0.396) | 220 (TP) | 498 (FN) | 0.306 (R) |
| | 437 (FP) | 19083 (TN) | 0.978 (TNR) |
| | 0.335 (P) | 0.975 (NPV) | 0.694 (FNR) |
| RF (Threshold 0.68) | 97 (TP) | 621 (FN) | 0.135 (R) |
| | 496 (FP) | 19024 (TN) | 0.975 (TNR) |
| | 0.164 (P)[1] | 0.968 (NPV) | 0.865 (FNR) |
| LR (Threshold 0.68) | 6 (TP) | 712 (FN) | 0.008 (R) |
| | 13 (FP) | 19507 (TN) | 0.999 (TNR) |
| | 0.316 (P)[1] | 0.965 (NPV) | 0.992 (FNR) |
| XGB (Threshold 0.55) | 146 (TP) | 572 (FN) | 0.203 (R) |
| | 691 (FP) | 18829 (TN) | 0.965 (TNR) |
| | 0.174 (P)[1] | 0.971 (NPV) | 0.797 (FNR) |

**Table 8.** Adjusting precision/recall performance to compare the models at a 20% precision level (1 true positive for every 4 false positives) for the within 72-hours BOS prediction window at $t_0$. Confusion matrices are depicted (TP, true positive; FP, false positive; TN, true negative; FN, false negative; P, precision; R, recall; TNR, true negative rate; NPV, negative predictive value; FNR, false negative rate). [1]Closest precision level gets to 20%.

| Model | Within 72-Hours BOS AKI at $t_3$ | | |
|---|---|---|---|
| CV-DBN (Threshold 0.32) | 101 (TP) | 617 (FN) | 0.141 (R) |
| | 405 (FP) | 19115 (TN) | 0.979 (TNR) |
| | 0.200 (P) | 0.969 (NPV) | 0.859 (FNR) |
| Chi²-DBN (Threshold 0.376) | 14 (TP) | 704 (FN) | 0.019 (R) |
| | 56 (FP) | 19464 (TN) | 0.997 (TNR) |
| | 0.200 (P) | 0.965 (NPV) | 0.981 (FNR) |
| IG-DBN (Threshold 0.318) | 129 (TP) | 589 (FN) | 0.180 (R) |
| | 517 (FP) | 19003 (TN) | 0.974 (TNR) |
| | 0.200 (P) | 0.970 (NPV) | 0.820 (FNR) |
| RF (Threshold 0.45) | 149 (TP) | 569 (FN) | 0.208 (R) |
| | 1187 (FP) | 18333 (TN) | 0.939 (TNR) |
| | 0.112 (P)[1] | 0.970 (NPV) | 0.792 (FNR) |
| LR (Threshold 0.46) | 88 (TP) | 630 (FN) | 0.123 (R) |
| | 347 (FP) | 19173 (TN) | 0.982 (TNR) |
| | 0.202 (P) | 0.968 (NPV) | 0.877 (FNR) |
| XGB (Threshold 0.347) | 228 (TP) | 490 (FN) | 0.318 (R) |
| | 1907 (FP) | 17613 (TN) | 0.902 (TNR) |
| | 0.107 (P)[1] | 0.973 (NPV) | 0.682 (FNR) |

**Supplementary Information**

**Automated Dynamic Bayesian Networks for Predicting Acute Kidney Injury Before Onset**

**David Gordon, MS[1,2,*], Panayiotis Petousis, PhD[3], Anders O. Garlid, PhD[2], Keith Norris, MD, PhD[4], Katherine Tuttle, MD[5], Susanne B. Nicholas, MD, MPH, PhD[4,+], and Alex A.T. Bui, PhD[1,2,3,+] on behalf of CURE-CKD**

[1]Department of Bioengineering, University of California, Los Angeles, CA, USA; [2]Medical & Imaging Informatics (MII) Group, Department of Radiological Sciences, University of California, Los Angeles, CA, USA; [3]UCLA Clinical and Translational Science Institute, Los Angeles, CA, USA; [4]Department of Medicine, David Geffen School of Medicine at UCLA, Los Angeles, CA, USA; [5]Providence Health Care, Spokane, WA, USA

[*]*Correspondence to: UCLA Medical Imaging Informatics, 924 Westwood Blvd., Suite 420, Los Angeles, CA 90024. Email-address: d.gordon@ucla.edu (D. Gordon)*
[+]*Senior authors had equal contributions*

**Supplementary Table S1.** BN AUCROC with 95% CIs and average precision-recall score.

| | Models | Metrics | $t_0$ |
|---|---|---|---|
| **Within 24-hours** | CV-BN | AUCROC CI | 0.821 [0.808, 0.834] |
| | | AP | 0.202 |
| | Chi$^2$-BN | AUCROC CI | 0.821 [0.808, 0.833] |
| | | AP | 0.201 |
| | IG-BN | AUCROC CI | 0.811 [0.797, 0.824] |
| | | AP | 0.204 |
| **Within 48-hours** | CV-BN | AUCROC CI | 0.711 [0.692, 0.732] |
| | | AP | 0.125 |
| | Chi$^2$-BN | AUCROC CI | 0.710 [0.693, 0.729] |
| | | AP | 0.090 |
| | IG-BN | AUCROC CI | 0.712 [0.693, 0.733] |
| | | AP | 0.129 |
| **Within 72-hours** | CV-BN | AUCROC CI | 0.698 [0.682, 0.715] |
| | | AP | 0.0982 |
| | Chi$^2$-BN | AUCROC CI | 0.708 [0.692, 0.724] |
| | | AP | 0.0977 |
| | IG-BN | AUCROC CI | 0.698 [0.682, 0.715] |
| | | AP | 0.0974 |

**Supplementary Table S2.** Top-BN-to-DBN AUCROC with 95% CIs and average precision-recall score.

|  | Models | Metrics | $t_0$ | $t_1$ | $t_2$ |
|---|---|---|---|---|---|
| **Within 24-hours** | IG-DBN | AUCROC CI | 0.811 [0.797, 0.824] | 0.758 [0.740, 0.778] | 0.825 [0.809, 0.841] |
|  |  | AP | 0.209 | 0.150 | 0.372 |
| **Within 48-hours** | IG-DBN | AUCROC CI | 0.718 [0.700, 0.739] | 0.731 [0.715, 0.748] | 0.786 [0.769, 0.804] |
|  |  | AP | 0.134 | 0.141 | 0.253 |
| **Within 72-hours** | CV-DBN | AUCROC CI | 0.700 [0.683, 0.717] | 0.732 [0.715, 0.749] | 0.757 [0.738, 0.774] |
|  |  | AP | 0.097 | 0.120 | 0.122 |
|  | Models | Metrics | $t_3$ | $t_4$ | $t_5$ |
| **Within 24-hours** | IG-DBN | AUCROC CI | 0.834 [0.816, 0.851] | 0.833 [0.814, 0.853] | 0.828 [0.806, 0.848] |
|  |  | AP | 0.406 | 0.378 | 0.363 |
| **Within 48-hours** | IG-DBN | AUCROC CI | 0.780 [0.760, 0.801] | 0.760 [0.736, 0.782] | x |
|  |  | AP | 0.249 | 0.225 | x |
| **Within 72-hours** | CV-DBN | AUCROC CI | 0.770 [0.752, 0.788] | x | x |
|  |  | AP | 0.124 | x | x |

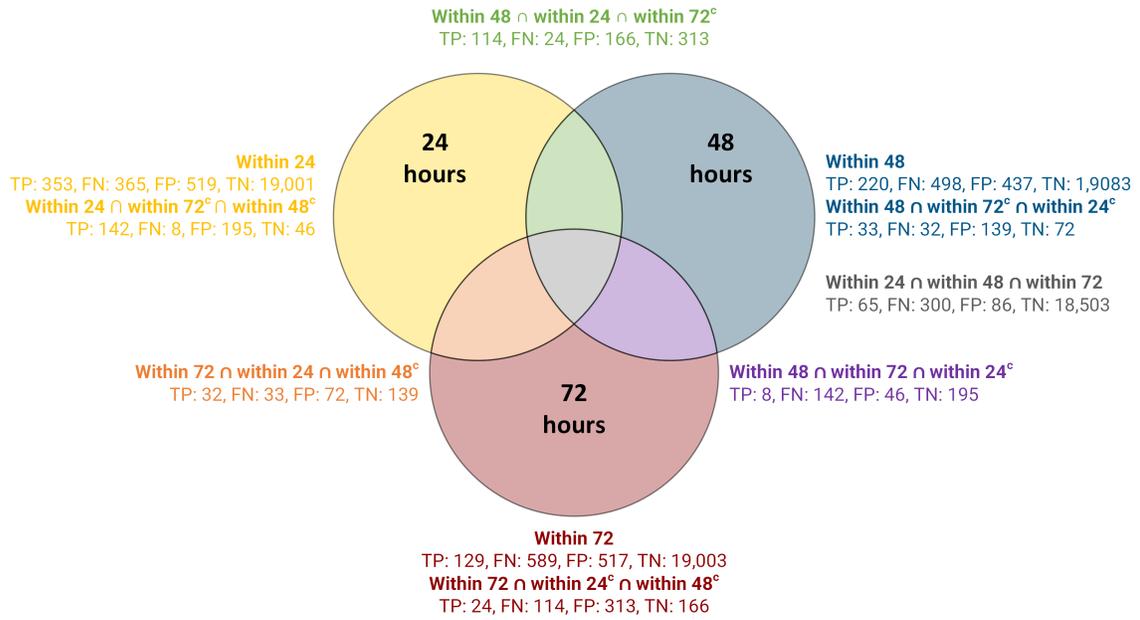

**Supplementary Figure S1.** Case agreement between the prediction windows for IG-DBN. Within 24 BOS at 40% precision level (yellow), within 48 BOS at 33% precision level (blue), and within 72 BOS at 20% precision level (red).

**Supplementary Table S3.** Summary of case agreement between the prediction windows for IG-DBN. Within 24 BOS at 40% precision level, within 48 BOS at 33% precision level, and within 72 BOS at 20% precision level.

| Intersections | Metrics | Within 24 | Within 48 | Within 72 |
|---|---|---|---|---|
| Within 24 BOS ∩ Within 48 BOS ∩ Within 72 BOS | TP | 18.4% | 29.5% | 50.4% |
|  | FN | 82.2% | 60.2% | 50.9% |
|  | FP | 16.6% | 19.7% | 16.6% |
| Within 24 BOS<sup>c</sup> ∩ Within 48 BOS ∩ Within 72 BOS | TP | x | 3.6% | 6.2% |
|  | FN | x | 28.5% | 24.1% |
|  | FP | x | 10.5% | 8.9% |
| Within 24 BOS ∩ Within 48 BOS<sup>c</sup> ∩ Within 72 BOS | TP | 9.1% | x | 24.8% |
|  | FN | 9.0% | x | 5.6% |
|  | FP | 13.9% | x | 13.9% |
| Within 24 BOS ∩ Within 48 BOS ∩ Within 72 BOS<sup>c</sup> | TP | 32.3% | 51.8% | x |
|  | FN | 6.6% | 4.8% | x |
|  | FP | 32.0% | 38.0% | x |
| None | TP | 40.0% | 15.0% | 18.6% |
|  | FN | 2.2% | 6.4% | 19.4% |
|  | FP | 37.6% | 31.8% | 60.5% |

**Supplementary Table S4.** TP, FN, and FP analyses for IG-DBN Within 24 BOS ∩ Within 48 BOS ∩ Within 72 BOS. I.e., temporal feature distribution between the prediction windows on testing data. Abbreviations: (BUN) blood urea nitrogen; (WBC) white blood cell count.

| | Features | Within 24 | | Within 48 | | Within 72 | |
|---|---|---|---|---|---|---|---|
| | | Effect Size | P-value | Effect Size | P-value | Effect Size | P-value |
| **TP (n=65)** | eGFR | 0.207 | **<0.001** | 0.183 | **<0.001** | 0.159 | **<0.001** |
| | Hemoglobin | 0.035 | 0.142 | 0.045 | 0.020 | 0.042 | 0.018 |
| | WBC | 0.042 | 0.054 | 0.041 | 0.042 | 0.056 | **≤0.001** |
| | Sodium | 0.014 | 0.855 | 0.031 | 0.210 | 0.025 | 0.327 |
| | BUN | 0.151 | **<0.001** | 0.134 | **<0.001** | 0.103 | **<0.001** |
| | Potassium | 0.041 | 0.156 | 0.048 | 0.037 | 0.022 | 0.556 |
| | Calcium | 0.059 | **<0.01** | 0.061 | **<0.01** | 0.047 | 0.017 |
| | Albumin | 0.063 | 0.105 | 0.077 | 0.018 | 0.082 | **<0.01** |
| **FN (n=300)** | eGFR | 0.153 | **<0.001** | 0.126 | **<0.001** | 0.147 | **<0.001** |
| | Hemoglobin | 0.040 | 0.064 | 0.043 | 0.029 | 0.036 | 0.066 |
| | WBC | 0.022 | 0.555 | 0.010 | 0.912 | 0.023 | 0.418 |
| | Sodium | 0.038 | 0.114 | 0.025 | 0.407 | 0.021 | 0.482 |
| | BUN | 0.074 | **<0.001** | 0.048 | 0.015 | 0.067 | **<0.001** |
| | Potassium | 0.055 | 0.024 | 0.049 | 0.028 | 0.078 | **<0.001** |
| | Calcium | 0.048 | 0.042 | 0.046 | 0.035 | 0.052 | **<0.01** |
| | Albumin | 0.117 | **<0.001** | 0.108 | **<0.001** | 0.089 | **<0.01** |
| **FP (n=86)** | eGFR | 0.204 | **<0.001** | 0.208 | **<0.001** | 0.191 | **<0.001** |
| | Hemoglobin | 0.037 | 0.113 | 0.034 | 0.128 | 0.048 | **<0.01** |
| | WBC | 0.019 | 0.651 | 0.037 | 0.090 | 0.042 | 0.022 |
| | Sodium | 0.018 | 0.719 | 0.007 | 0.967 | 0.023 | 0.437 |
| | BUN | 0.165 | **<0.001** | 0.166 | **<0.001** | 0.134 | **<0.001** |
| | Potassium | 0.016 | 0.859 | 0.034 | 0.236 | 0.029 | 0.302 |
| | Calcium | 0.056 | **≤0.01** | 0.064 | **≤0.001** | 0.067 | **<0.001** |
| | Albumin | 0.089 | **<0.01** | 0.084 | **<0.01** | 0.091 | **≤0.001** |

**Supplementary Table S5.** TP, FN, and FP analyses for IG-DBN Within 24 BOS ∩ Within 48 BOS ∩ Within 72 BOS. I.e., static feature distribution on testing data.

| Metric | Static Features | Effect Size | P-value |
|---|---|---|---|
| TP (n=65) | Age | 0.018 | 0.229 |
| | Gender | 0.006 | 0.391 |
| | Race | 0.022 | 0.864 |
| | Hypertension | 0.031 | **<0.001** |
| | Diabetes | 0.015 | 0.034 |
| FN (n=300) | Age | 0.061 | **<0.001** |
| | Gender | 0.025 | **<0.001** |
| | Race | 0.052 | **<0.001** |
| | Hypertension | 0.028 | **<0.001** |
| | Diabetes | 0.006 | 0.421 |
| FP (n=86) | Age | 0.030 | **<0.01** |
| | Gender | 0.011 | 0.134 |
| | Race | 0.014 | 0.998 |
| | Hypertension | 0.025 | **<0.001** |
| | Diabetes | 0.009 | 0.201 |

**Supplementary Table S6**. Early true positives (ETP) for IG-DBN at Within 48-hours BOS and Within 72-hours BOS.

|  | Percent |
|---|---|
| Within 48-hours ETP for Within 24-hours | 34.9 |
| Within 72-hours ETP for Within 48-hours | 25.0 |
| Within 72-hours ETP for Within 24-hours | 43.5 |